\documentclass[runningheads]{llncs}
\usepackage[T1]{fontenc}
\usepackage{graphicx,verbatim}
\usepackage{booktabs}
\usepackage{amsmath,amssymb,amsfonts}
\usepackage{subcaption}
\usepackage{multirow}
\usepackage{bbm}
\usepackage{bbding}

\usepackage{hyperref}
\usepackage{color}

\begin{document}

\title{Radiologist Copilot: An Agentic Framework Orchestrating Specialized Tools for Reliable Radiology Reporting}
\titlerunning{Radiologist Copilot}

\author{Yongrui Yu\inst{1}, Zhongzhen Huang\inst{1}, Linjie Mu\inst{1}, Shaoting Zhang\inst{1,2}\Envelope, and Xiaofan Zhang\inst{1,3}\Envelope}
\authorrunning{Y. Yu et al.}
\institute{Shanghai Jiao Tong University, Shanghai, China \and SenseTime Research, Shanghai, China \and Shanghai Innovation Institute, Shanghai, China \\
\email{zhangshaoting@sensetime.com; xiaofan.zhang@sjtu.edu.cn}}

\maketitle

\begin{abstract}
In clinical practice, radiology reporting is an essential yet complex, time-intensive, and error-prone task, particularly for 3D medical images. Existing automated approaches based on medical vision-language models primarily focus on isolated report generation. However, real-world radiology reporting extends far beyond report writing, which requires meticulous image observation and interpretation, appropriate template selection, and rigorous quality control to ensure adherence to clinical standards. This multi-stage, planning-intensive workflow fundamentally exceeds the capabilities of single-pass models. To bridge this gap, we propose Radiologist Copilot, an agentic system that autonomously orchestrates specialized tools to complete the entire radiology reporting workflow rather than isolated report writing. Radiologist Copilot enables region image localization and region analysis planning to support detailed visual reasoning, adopts strategic template selection for standardized report writing, and incorporates dedicated report quality control via quality assessment and feedback-driven iterative refinement. By integrating localization, interpretation, template selection, report composition, and quality control, Radiologist Copilot delivers a comprehensive and clinically aligned radiology reporting workflow. Experimental results demonstrate that it significantly outperforms state-of-the-art methods, supporting radiologists throughout the entire radiology reporting process. The code will be released upon acceptance.
\keywords{Radiology Reporting \and Medical Agents \and Specialized Tools.}
\end{abstract}

\section{Introduction}
Radiology reporting plays a crucial role in radiological examinations, requiring radiologists to carefully interpret medical images and produce comprehensive reports that summarize key findings and impressions. This process is often complex, time-consuming, and error-prone, particularly for 3D medical imaging modalities~\cite{hamamci2024ct2rep}. Existing automated methods for volumetric medical images, such as CT2Rep~\cite{hamamci2024ct2rep}, medical vision-language models (VLMs)~\cite{wu2025towards,bai2024m3d,hamamci2024developing,jiang2025hulu}, and more recently agentic models~\cite{mao2025ct,fallahpour2025medrax}, have primarily focused on isolated report generation. However, in practice, radiology reporting extends far beyond report generation, involving region-of-interest image localization, detailed image analysis, suitable template selection, and rigorous quality control to ensure that the report is accurate, complete, and free of errors~\cite{warr2024quality}. Such a multi-stage, planning-intensive workflow is inherently beyond the reach of single-pass methods.

To address this challenge, \textit{we adopt an agentic approach that enables autonomous planning and execution of multi-step actions, handling the complex radiology reporting workflow, from image analysis and template selection to report composition and quality assurance}. We introduce Radiologist Copilot, an agentic AI framework that leverages large language models (LLMs) as its reasoning backbone and orchestrates specialized tools to support the complete radiology reporting workflow. Beyond isolated report generation, Radiologist Copilot incorporates region-of-interest image localization and region analysis planning to highlight and examine critical anatomical structures within the region images. To facilitate standardized report writing, Radiologist Copilot implements strategic template selection, enabling the framework to choose appropriate templates. Additionally, report quality control is integrated to ensure that final reports meet clinical standards through automated quality assessment and feedback-driven iterative refinement. Rather than being limited to isolated report generation, Radiologist Copilot delivers a comprehensive, clinically aligned radiology reporting workflow, assisting radiologists and improving clinical efficiency. The final radiology report not only contains detailed findings and impression sections, but also provides key slice references of the volumetric images, offering both textual and imaging evidence that makes the report traceable and well-founded.

In conclusion, our main contributions are summarized as follows:
\begin{itemize}
\item We propose \textit{Radiologist Copilot}, an agentic framework that autonomously orchestrates specialized tools to tackle the complex radiology reporting process, from image analysis and template selection to report composition and quality assurance, aligning the workflow more closely with clinical practice.
\item We introduce \textit{region analysis planning} to identify and examine critical structures within region-of-interest images, enabling more meticulous visual observation and interpretation. To facilitate standardized report writing, we utilize \textit{strategic template selection} to choose the most appropriate templates. We further introduce \textit{report quality control} via quality assessment and feedback-driven adaptive refinement to ensure adherence to clinical standards.
\item Experimental results show that Radiologist Copilot surpasses state-of-the-art methods and facilitates a complete and reliable automated radiology reporting process.
\end{itemize}

\section{Related Work}
Automated radiology reporting is crucial for supporting radiologists, since the process for 3D medical imaging modalities is extremely challenging. Previous methods often focus on report generation, which constitutes only one stage of the entire radiology reporting process and lacks critical procedures such as template selection and quality control.
CT2Rep~\cite{hamamci2024ct2rep} utilizes an auto-regressive causal transformer architecture and relational memory for report generation of 3D chest CT images. Besides, Reg2RG~\cite{chen2025large} leverages a region-guided referring and grounding framework for chest CT report generation. 

Medical VLMs can also be applied to report generation. In particular, 3D medical VLMs have shown effectiveness for 3D medical modalities.
RadFM~\cite{wu2025towards} introduces a generalist vision-language foundation model that unifies both 2D and 3D medical data for radiology.
M3D~\cite{bai2024m3d} presents M3D-Data, a large-scale multi-modal medical dataset, together with M3D-LaMed, a multi-modal LLM for 3D medical image analysis.
Merlin~\cite{blankemeier2024merlin} develops a 3D VLM that leverages both structured and unstructured data for abdominal CT interpretation.
CT-CHAT~\cite{hamamci2024developing} is a vision-language foundation model for 3D chest CT volumes, combining the CT-CLIP vision encoder with a pretrained LLM.
Med3DVLM~\cite{xin2025med3dvlm} is a 3D VLM that enhances multi-modal representations via an efficient encoder, a contrastive learning strategy, and a dual-stream projector.
Hulu-Med~\cite{jiang2025hulu} provides a generalist medical VLM that unifies text, 2D and 3D images, and video within a single architecture.

The advent of AI agents enables autonomous reasoning, planning, and tool usage. Recently, medical agents have been introduced for complex medical tasks. For example, MMedAgent~\cite{li2024mmedagent} presents a multi-modal medical agent that leverages various tools to handle medical tasks across different modalities, such as radiology report generation for chest X-rays (CXR). MedRAX~\cite{fallahpour2025medrax} provides a specialized agent framework for chest X-ray interpretation that integrates state-of-the-art CXR analysis tools. CT-Agent~\cite{mao2025ct} is an anatomy-aware and token-efficient agent for 3D CT visual question answering, enabling effective reasoning. However, for the report generation task, these agentic methods typically rely on an isolated tool and lack coordination among tools. In contrast, our Radiologist Copilot provides diverse and comprehensive tools, facilitates inter-tool collaboration, and supports more sophisticated tool planning and execution.

\begin{figure}[!t]
\centering
\includegraphics[page=1, trim={9cm, 7cm, 11.5cm, 4cm}, clip, width=0.6\textwidth]{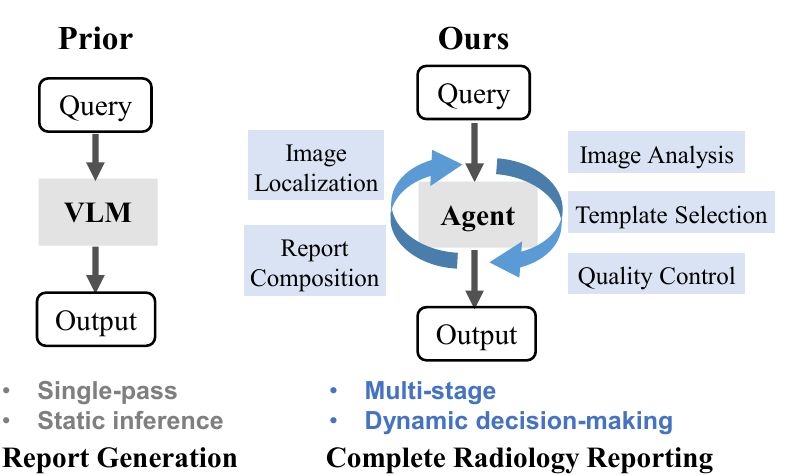}
\caption{Comparison with prior work.}
\label{fig:compare}
\end{figure}

\section{Methodology}
While existing approaches primarily focus on isolated report generation, Radiologist Copilot extends far beyond this scope. Complete radiology reporting is inherently a multi-stage, context-dependent, and complex decision-making process. To address this challenge, we adopt an agentic framework that leverages the reasoning capabilities of LLMs to perform autonomous multi-step planning and execution, enabling coordinated collaboration across diverse specialized tools. Radiologist Copilot thus delivers a clinically aligned radiology reporting workflow, supporting radiologists and enhancing clinical efficiency.

\begin{figure}[!t]
\centering
\includegraphics[page=2, trim={5cm, 12.5cm, 5cm, 0cm}, clip, width=\textwidth]{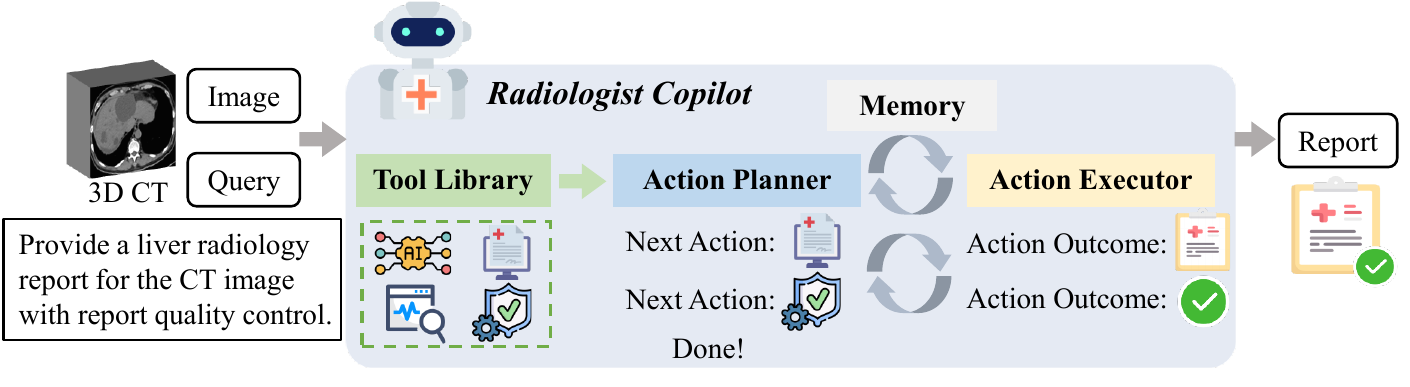}
\caption{Overview of Radiologist Copilot, an agentic framework autonomously orchestrating specialized tools to enable complete and reliable radiology reporting.}
\label{fig:agent}
\end{figure}

\paragraph{\textbf{Radiologist Copilot.}}
As shown in Fig.~\ref{fig:agent}, Radiologist Copilot is proposed to complete the complex radiology reporting process via multi-step planning and execution. Given an input query $Q$ and a 3D CT image $I\in\mathbb{R}^{C\times H\times W\times D}$, Radiologist Copilot produces a qualified report $R_{\text{qual}}$. The agentic assistant uses $\text{LLM}_{\theta_l}(\cdot)$ as its reasoning backbone and employs an action planner and an action executor to autonomously take actions. The action planner first formulates a high-level solution to guide subsequent actions and then determines the next action by selecting the appropriate tool. The action executor generates and executes commands for the next action and obtains action outcomes. A memory module records each action and its outcomes, enabling verification of task completion. The dynamic decision-making process of the agentic system relies on the environment state, observations, and historical trajectories, with a subsequent action $a_t$ sampled from the LLM policy $\pi_{\theta_l}$ based on the previous history $h_{t-1}$; action $a_t$ triggers a state transition to $s_t$ and produces a new observation $o_t$.

\begin{equation}
a_t\sim\pi_{\theta_l}(a\mid h_{t-1}),\;(s_t,o_t)=\mathcal{T}(s_{t-1},a_t),\;h_t=h_{t-1}\cup\{a_t,o_t\}
\label{eq:agent}
\end{equation}

\begin{figure}[!t]
\centering
\includegraphics[page=2, trim={3.5cm, 2cm, 3cm, 6.5cm}, clip, width=\textwidth]{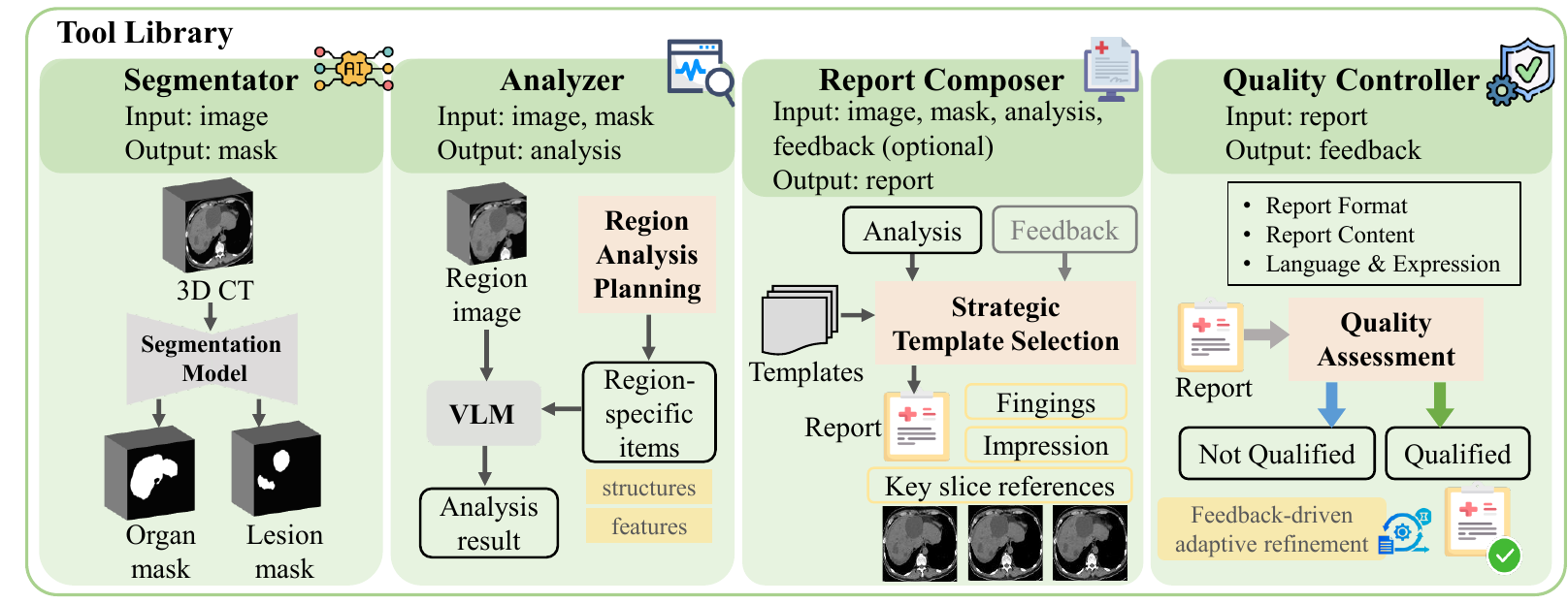}
\caption{The specialized tool library equipped with region analysis planning, strategic template selection, and report quality control.}
\label{fig:tool}
\end{figure}

We curate a suite of specialized tools into a tool library $\{T_i\}_{i=1}^{n}$ as shown in Fig.~\ref{fig:tool} to provide essential capabilities for automated radiology reporting. We introduce region-of-interest image localization in the Segmentator Tool and region analysis planning in the Analyzer Tool. In addition, we incorporate strategic template selection into the Report Composer Tool for standardized report writing. The Quality Controller Tool implements report quality control, ensuring that the radiology reports meet clinical standards through quality assessment and feedback-driven adaptive refinement. The agentic assistant orchestrates these specialized tools to facilitate complete and reliable radiology reporting.

\paragraph{\textbf{Region Analysis Planning.}}
To enable meticulous image observation and interpretation, the region-of-interest image is localized using pretrained segmentation models~\cite{wasserthal2023totalsegmentator} to obtain query-relevant organ mask $M_{\text{organ}}$ and lesion mask $M_{\text{lesion}}$. The region image $I_{\text{region}}$ is then extracted from $I$ based on $M_{\text{organ}}$. We propose region analysis planning to emphasize critical anatomical structures within the region-of-interest image, enabling detailed, region-specific image investigation. The region analysis planning identifies region-specific analysis items that encompass anatomical structures and clinically significant characteristics, such as size, shape, and density. In addition, region analysis planning adaptively determines whether lesion-related characteristics need to be analyzed, depending on the presence of lesions in $M_{\text{lesion}}$.

\begin{equation}
\mathcal{I}=\mathcal{I}_{\text{organ}}\cup\mathbbm{1}(|M_{\text{lesion}}|>0)\cdot\mathcal{I}_{\text{lesion}}=\{i_k\}_{k=1}^{K},\;i_k\in\{\text{structure},\text{feature}\}
\label{eq:rap}
\end{equation}

These analysis items $\mathcal{I}$ are formulated as prompts $\phi(\mathcal{I})$ for 3D medical VLMs to generate region analysis results $Ana=\text{VLM}_{\theta_v}(I_{\text{region}},\phi(\mathcal{I}))$.

\paragraph{\textbf{Strategic Template Selection.}}
Template selection is often ignored, but crucial for structured and standardized radiology report writing. To facilitate appropriate template selection, we introduce strategic template selection, where templates are selected based on analysis results. Candidate template reports $\mathcal{C}$ are obtained via unsupervised clustering of historical radiology reports. Given these template reports, the LLM selects the most relevant template based on current analysis. The report writing leverages the selected template as a reference while adapting to current analysis. Furthermore, strategic template selection and report writing can incorporate \textbf{feedback comments} from the Quality Controller Tool, enabling iterative revision of reports that do not meet quality standards.

\begin{equation}
C^*=\arg\max_{C_j\in\mathcal{C}}p(C_j\mid Ana,\;feedback\text{(opt)})
\label{eq:sts}
\end{equation}

\begin{equation}
R_{\text{gen}}=\text{LLM}_{\theta_l}(Ana,\;feedback\text{(opt)},\;C^*)
\label{eq:rw}
\end{equation}

Since formal radiology reports often reference CT slices, we select the central three axial slices: from the organ for normal cases and from the largest lesion for abnormal cases. The generated report $R_{\text{gen}}$ comprises three components: the findings section, the impression section, and the key slice references, providing both textual and imaging evidence to enhance traceability.
\begin{equation}
R_{\text{gen}} = R_{\text{findings}} \cup R_{\text{impression}} \cup R_{\text{keyslices}}
\label{eq:report}
\end{equation}

\paragraph{\textbf{Report Quality Control.}}
Quality control is critical to ensure that radiology reports adhere to clinical standards, but is often overlooked in previous studies. The Quality Controller Tool performs a \textbf{quality assessment} (QA) on the generated report $R_{\text{gen}}$ and produces \textbf{feedback comments} using the LLM. The quality assessment conducts a thorough evaluation of the generated report, covering format, content, language, and expression. Specifically, the report format is assessed to ensure that the findings provide an objective description of imaging manifestations, while the impression presents a diagnostic summary emphasizing key conclusions. The content is evaluated for anatomical correctness, lesion characterization, and consistency. Language and expression are examined to ensure standardized radiological terminology, correct spelling, clarity, and conciseness. The feedback comments reflect whether the report is qualified. If the report is considered qualified, i.e., $\text{QA}(R_{\text{gen}})=1$, it is regarded as qualified report $R_{\text{qual}}$. Otherwise, i.e., $\text{QA}(R_{\text{gen}})=0$, the feedback comments identify deficiencies and are used to enable \textbf{feedback-driven adaptive refinement} of $R_{\text{gen}}$.

\begin{equation}
R_{\text{gen}}^{(t+1)}=\text{Adaptive Refinement}\big(R_{\text{gen}}^{(t)},\;feedback(R_{\text{gen}}^{(t)})\big),\;t=0,1,2,\dots
\label{rqc}
\end{equation}

Finally, by integrating localization, analysis, template selection, report composition, and quality control, Radiologist Copilot surpasses isolated report generation, delivering a complete and clinically aligned radiology reporting workflow.

\section{Experiments and Results}
\subsection{Experimental Setup}
\paragraph{\textbf{Datasets.}}
We leverage liver radiology reporting task to validate the effectiveness of our proposed Radiologist Copilot. We utilize the publicly available 3D CT dataset AMOS-MM~\cite{ji2022amos} and its medical report generation task, which contains 1,287 training cases and 400 validation cases. For the liver radiology reporting task, we filter the reports in the AMOS-MM dataset to identify those containing liver descriptions and extract these descriptions as liver radiology reports. The resulting training set consists of 1,149 CT scans with liver reports, and the validation set contains 367 CT scans with liver reports. The validation set is reserved, while the training set is used to conduct liver report clustering. We first obtain report embeddings using BioBERT~\cite{lee2020biobert}, and then perform K-means clustering on these embeddings to derive template reports. These template reports are summarized into liver analysis items, including liver surface, liver parenchyma, bile ducts, and liver lesions, which are used in the region analysis planning.

\paragraph{\textbf{Implementation Details.}}
Radiologist Copilot is implemented based on the agent framework OctoTools~\cite{lu2025octotools}, and experiments are conducted on NVIDIA L20 GPUs. The agentic framework operates in a training-free manner using pretrained models. We utilize Qwen3-32B~\cite{yang2025qwen3} as the LLM backbone for agent reasoning. The Segmentator Tool employs the TotalSegmentator~\cite{wasserthal2023totalsegmentator} as the segmentation model, while the Analyzer Tool leverages the 3D medical VLM Hulu-Med~\cite{jiang2025hulu} for CT analysis. The maximum number of steps is set to 10.

\paragraph{\textbf{Evaluation Metrics.}}
To thoroughly evaluate the effectiveness of Radiologist Copilot, we perform both agent-level and task-level evaluations.
For task-level evaluation of radiology reporting, we adopt natural language generation (NLG) and clinical efficacy (CE) metrics. NLG metrics include BLEU-1~\cite{papineni2002bleu}, ROUGE-L~\cite{lin2004rouge}, METEOR~\cite{banerjee2005meteor}, and BERTScore~\cite{zhang2019bertscore}. CE metrics contain F1-RadGraph~\cite{delbrouck2024radgraph} and GREEN~\cite{ostmeier2024green}.
For agent-level evaluation, we use LLM-as-a-Judge~\cite{zheng2023judging}, employing OpenAI GPT-5.2~\cite{openai2025gpt52} as the judge to assess the entire automated radiology reporting process across four dimensions (Problem Analysis, Action Planning, Action Execution, and Overall Workflow), with scores ranging from 1 to 5 (Poor, Fair, Moderate, Good, Excellent).
Problem Analysis evaluates whether the agent's task understanding is comprehensive and logical, and whether the overall solution is coherent and reasonable. Action Planning assesses whether the agent selects suitable tools and plans its actions reasonably. Action Execution examines whether the agent successfully executes the planned actions, using appropriate commands and producing valid results. Overall Workflow measures the effectiveness, reliability, and coherence of the agent’s complete solution process.

\subsection{Experimental Results}
To comprehensively evaluate the effectiveness of Radiologist Copilot, we perform both agent-level evaluations of the agentic framework and task-level evaluations of liver radiology reporting, along with ablation studies on its core components.

\begin{figure}[!t]
\centering
\includegraphics[page=1, trim={0cm, 8.5cm, 0cm, 4cm}, clip, width=0.9\textwidth]{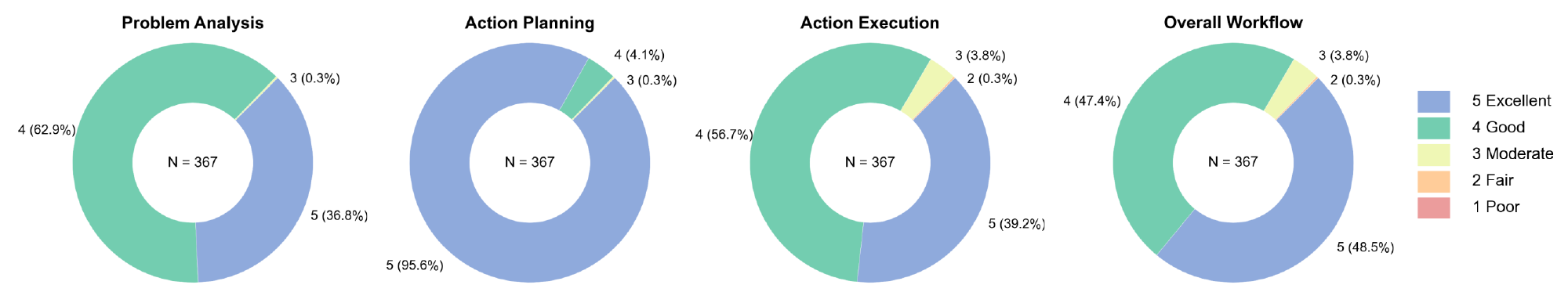}
\caption{Agent-level evaluation of Radiologist Copilot using LLM-as-a-Judge.}
\label{fig:evaluation}
\end{figure}

\begin{figure}[!t]
\centering
\begin{subfigure}[b]{0.49\textwidth}
\centering
\includegraphics[page=3, trim={7.5cm, 1.5cm, 7.5cm, 0cm}, clip, width=1.0\textwidth]{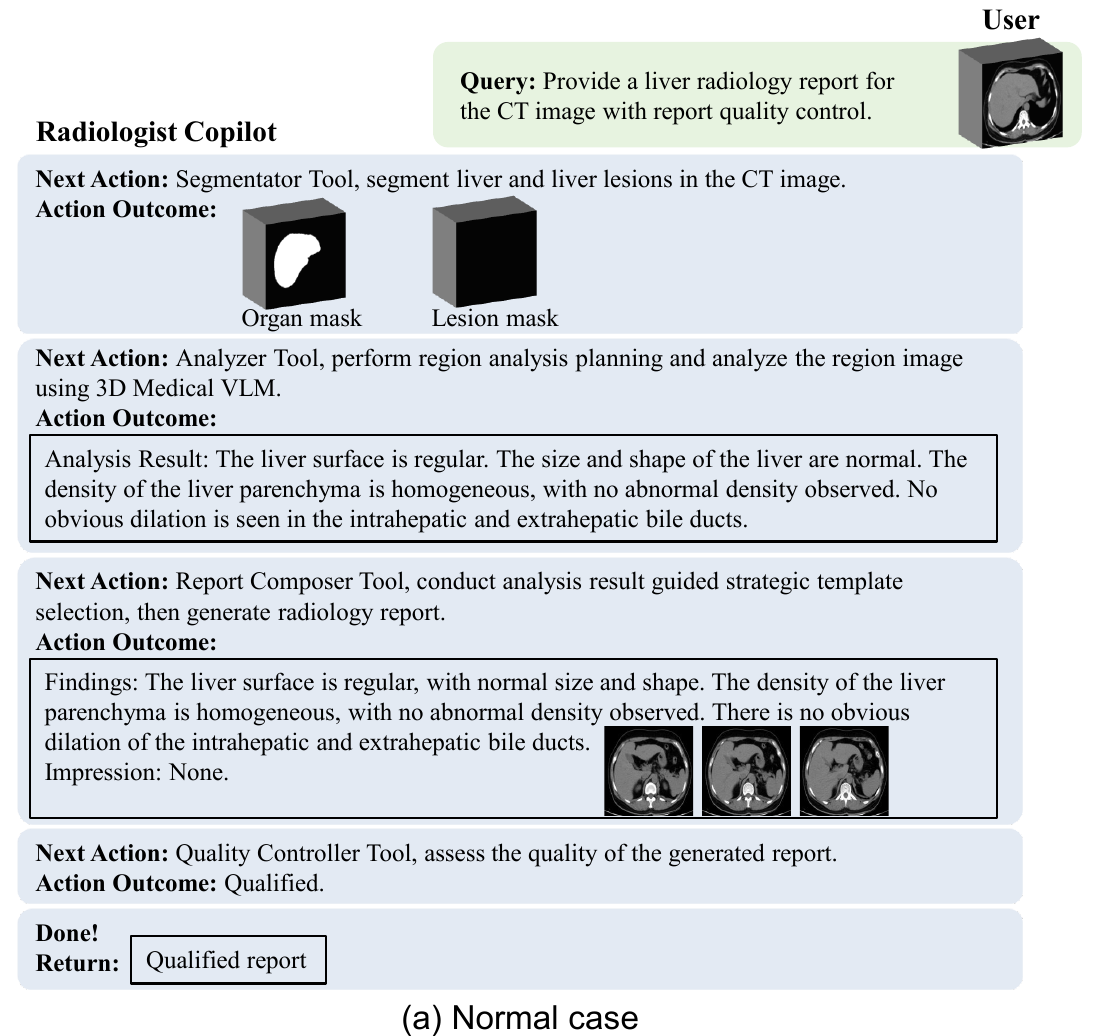}
\end{subfigure}
\begin{subfigure}[b]{0.49\textwidth}
\centering
\includegraphics[page=4, trim={7.5cm, 0cm, 7.5cm, 0cm}, clip, width=1.0\textwidth]{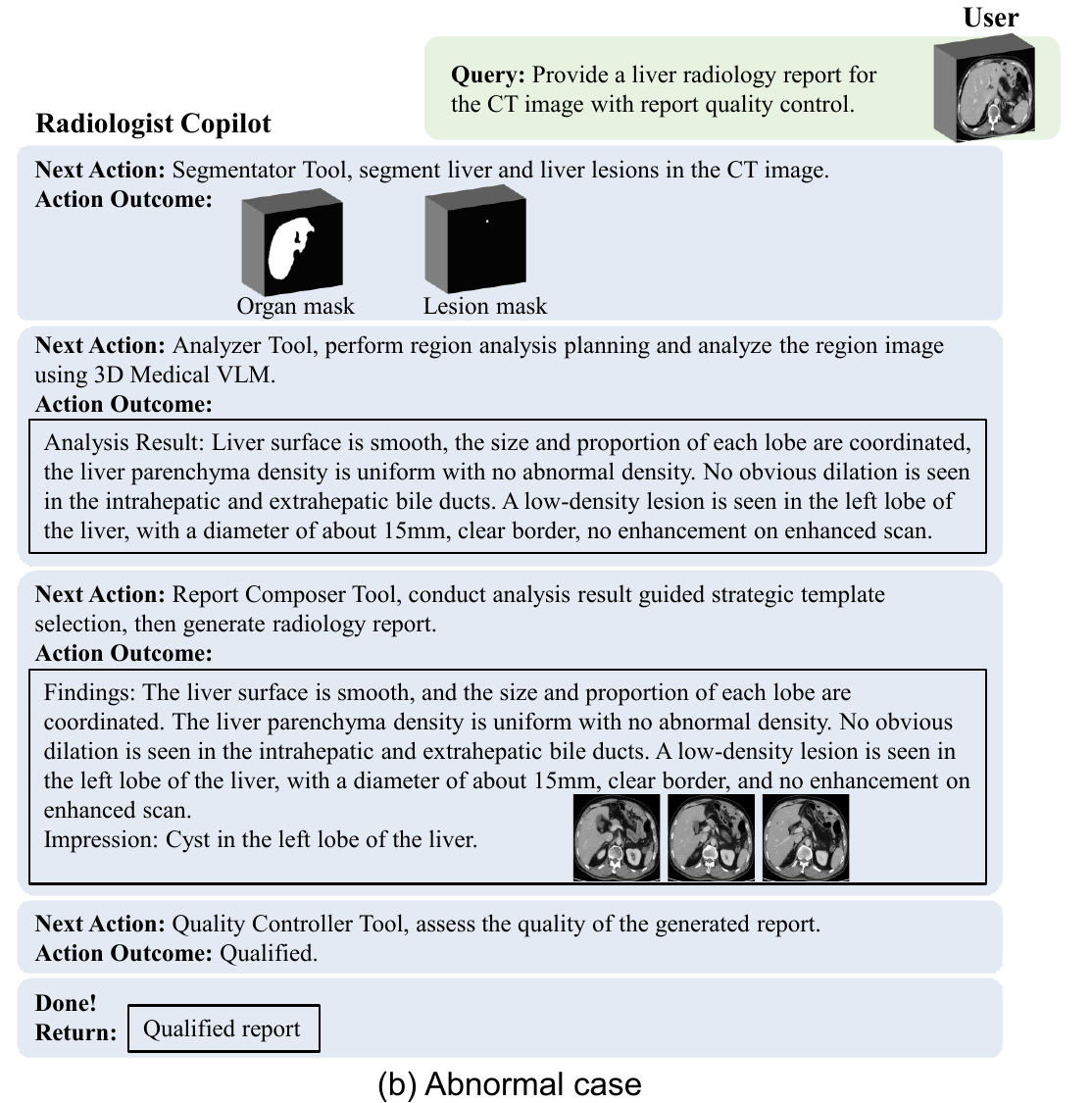}
\end{subfigure}
\caption{Examples of the radiology reporting workflow using Radiologist Copilot.}
\label{fig:example}
\end{figure}

\paragraph{\textbf{Agent-level Evaluation.}}
Fig.~\ref{fig:evaluation} presents LLM-as-a-Judge assessments on the entire radiology reporting process of Radiologist Copilot. This figure shows the score distribution across four dimensions, with the majority scores being 5 (Excellent) or 4 (Good), and only a small portion being 3 (Moderate). These results demonstrate that the agent’s Problem Analysis, Action Planning, Action Execution and Overall Workflow are effective and reliable. Furthermore, Fig.~\ref{fig:example} illustrates the radiology reporting workflow of Radiologist Copilot, demonstrating that action planning and action execution are coherent and efficient, ultimately resulting in a qualified radiology report. The entire process emulates radiologists’ practices in radiology reporting and enhances clinical efficiency.

\begin{figure}[!t]
\centering
\begin{subfigure}[b]{0.75\textwidth}
\centering
\includegraphics[page=2, trim={3cm, 2.5cm, 2cm, 1.5cm}, clip, width=1.0\textwidth]{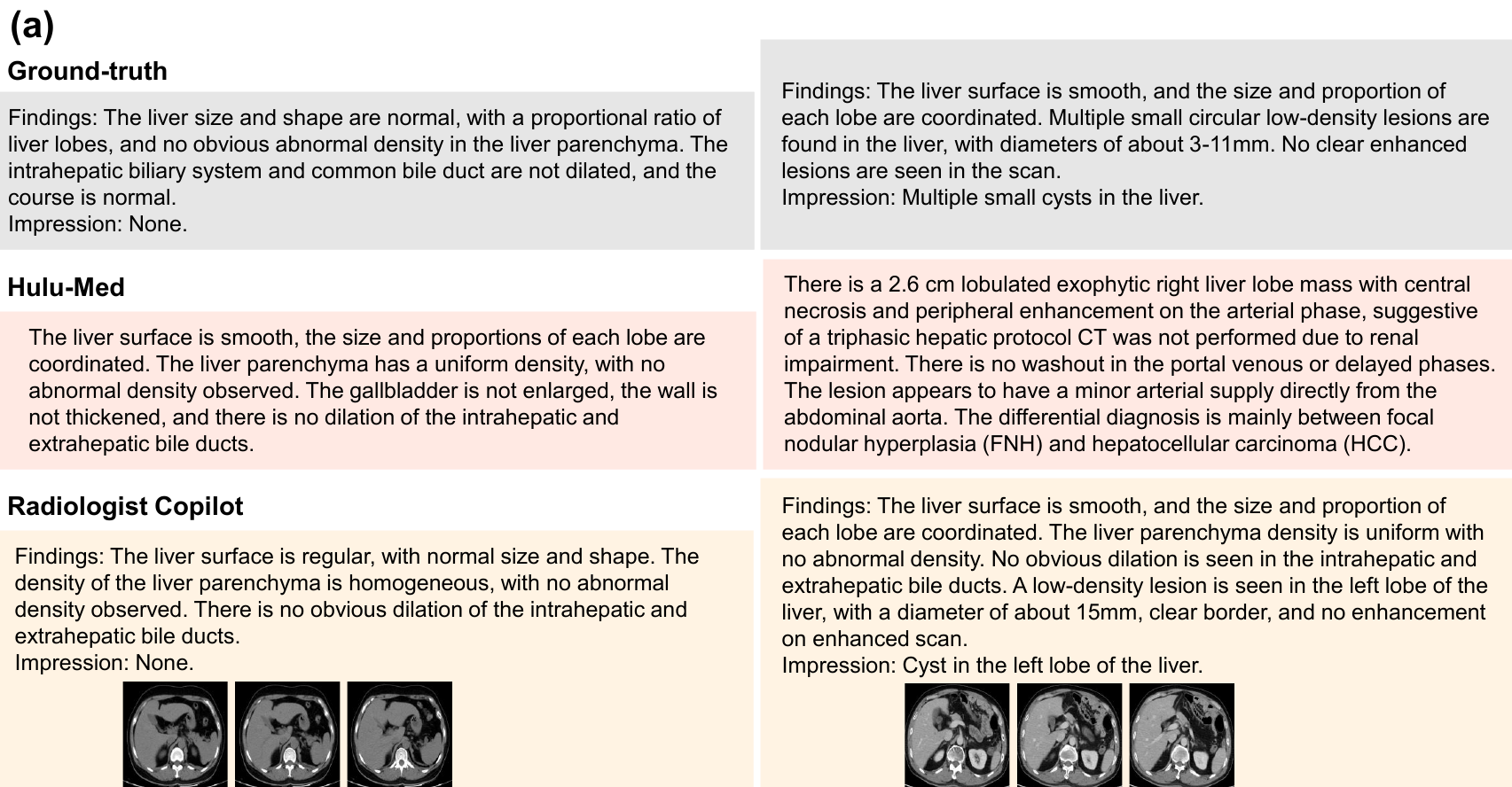}
\end{subfigure}
\begin{subfigure}[b]{0.23\textwidth}
\centering
\includegraphics[page=5, trim={8cm, 4.5cm, 16cm, 1.5cm}, clip, width=1.0\textwidth]{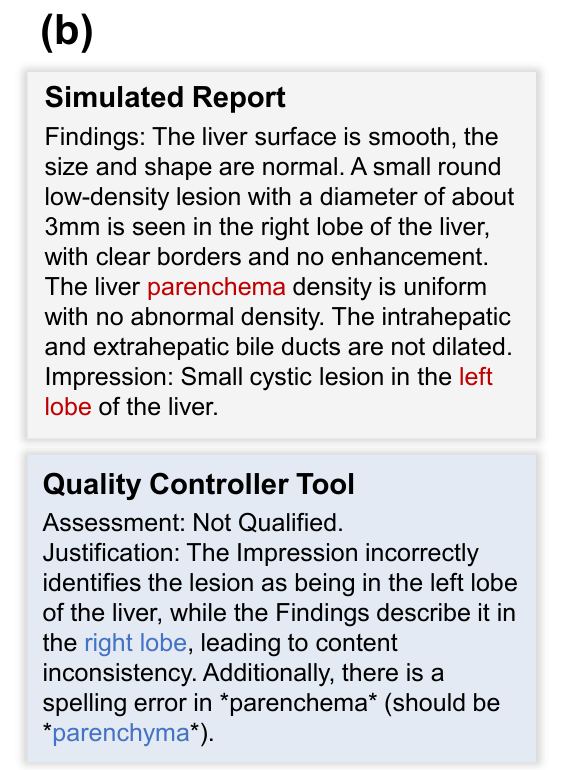}
\end{subfigure}
\caption{(a) Case study. (b) Validation of the report quality control.}
\label{fig:result}
\end{figure}

\paragraph{\textbf{Task-level Evaluation.}}
We compare the liver radiology reporting performance of our proposed method with state-of-the-art report generation approaches. As shown in Table~\ref{tab:comparison}, Radiologist Copilot substantially outperforms 3D medical vision-language models on both NLG and CE metrics, demonstrating its practical effectiveness in radiology reporting. Fig.~\ref{fig:result}(a) presents generated liver reports, which show that the reports produced by Radiologist Copilot are highly consistent with the ground-truth in both format and content, significantly surpassing the performance of Hulu-Med~\cite{jiang2025hulu}. These results indicate that reports produced via the complete radiology reporting workflow of our proposed agentic framework are superior to those produced by methods limited to isolated report generation.

\begin{table}[!t]
\centering
\caption{Task-level evaluation on the liver radiology reporting task.}
\begin{tabular}{p{2.5cm}|cccc|cc}
\toprule
Method & BLEU-1 & ROUGE-L & METEOR & BERTScore & F1-RadGraph & GREEN \\
\midrule
\multicolumn{7}{l}{\textbf{3D Medical VLM}} \\
RadFM~\cite{wu2025towards} & 0.1492 & 0.1415 & 0.2340 & 0.5541 & 0.0686 & 0.0353 \\
M3D~\cite{bai2024m3d} & 0.1775 & 0.1302 & 0.1220 & 0.5359 & 0.0475 & 0.0209 \\
Merlin~\cite{blankemeier2024merlin} & 0.0015 & 0.0908 & 0.0569 & 0.5119 & 0.1617 & 0.1024 \\
CT-CHAT~\cite{hamamci2024developing} & 0.2440 & 0.2012 & 0.2599 & 0.6127 & 0.1196 & 0.0390 \\
Med3DVLM~\cite{xin2025med3dvlm} & 0.1967 & 0.1422 & 0.1847 & 0.5608 & 0.0660 & 0.0539 \\
Hulu-Med~\cite{jiang2025hulu} & 0.1867 & 0.1723 & 0.2380 & 0.5947 & 0.1209 & 0.2163 \\
\midrule
\multicolumn{7}{l}{\textbf{3D Medical Agent}} \\
Ours & \textbf{0.4025} & \textbf{0.3222} & \textbf{0.4560} & \textbf{0.7024} & \textbf{0.2585} & \textbf{0.4379} \\
\bottomrule
\end{tabular}
\label{tab:comparison}
\end{table}

\paragraph{\textbf{Ablation Studies.}}
We perform ablation studies on the key components of Radiologist Copilot, with results presented in Table~\ref{tab:ablation}. The results underscore the importance of region analysis planning and strategic template selection, as removing either component leads to a noticeable degradation in the report quality. Although removing strategic template selection slightly improves the GREEN metric, other metrics decrease significantly. This component offers standardized templates and ensures structural consistency. From the results in the table, the impact of report quality control is less prominent compared to other components, as reports generated throughout the entire process of Radiologist Copilot generally meet the required standards. Nevertheless, report quality control remains important, providing the assurance that generated reports adhere to clinical standards. To illustrate its effectiveness, we provide a separate assessment of the Quality Controller Tool in Fig.~\ref{fig:result}(b), which effectively detects issues in the simulated report, including content inconsistencies and spelling errors.
In addition, we conduct an ablation study on Radiologist Copilot equipped with different 3D medical VLMs to perform region-of-interest image analysis. The results in Table~\ref{tab:vlm} indicate that even with different VLMs, Radiologist Copilot consistently maintains robust radiology reporting performance, demonstrating the superiority of our proposed agentic framework with specialized tools.

\begin{table}[!t]
\centering
\caption{Ablation study on the components of our proposed Radiologist Copilot. RAP, STS, and RQC denote region analysis planning, strategic template selection, and report quality control, respectively.}
\begin{tabular}{l|cccc|cc}
\toprule
Method & BLEU-1 & ROUGE-L & METEOR & BERTScore & F1-RadGraph & GREEN \\
\midrule
Ours & \textbf{0.4025} & \textbf{0.3222} & \textbf{0.4560} & \textbf{0.7024} & \textbf{0.2585} & 0.4379 \\
Ours (w/o RAP) & 0.3600 & 0.2588 & 0.3610 & 0.6469 & 0.1675 & 0.2269 \\
Ours (w/o STS) & 0.2983 & 0.2698 & 0.3915 & 0.6553 & 0.2429 & \textbf{0.4582} \\
Ours (w/o RQC) & 0.3998 & 0.3149 & 0.4462 & 0.6944 & 0.2545 & 0.4360 \\
\bottomrule
\end{tabular}
\label{tab:ablation}
\end{table}

\begin{table}[!t]
\centering
\caption{Ablation study on Radiologist Copilot equipped with different VLMs.}
\begin{tabular}{l|cccc|cc}
\toprule
Method & BLEU-1 & ROUGE-L & METEOR & BERTScore & F1-RadGraph & GREEN \\
\midrule
RadFM & 0.1492 & 0.1415 & 0.2340 & 0.5541 & 0.0686 & 0.0353 \\
Ours (RadFM) & 0.3215 & 0.2465 & 0.3633 & 0.6510 & 0.1994 & 0.3719 \\
\midrule
CT-CHAT & 0.2440 & 0.2012 & 0.2599 & 0.6127 & 0.1196 & 0.0390 \\
Ours (CT-CHAT) & 0.3671 & 0.2784 & 0.3738 & 0.6558 & 0.2312 & 0.1485 \\
\midrule
Hulu-Med & 0.1867 & 0.1723 & 0.2380 & 0.5947 & 0.1209 & 0.2163 \\
Ours (Hulu-Med) & 0.4025 & 0.3222 & 0.4560 & 0.7024 & 0.2585 & 0.4379 \\
\bottomrule
\end{tabular}
\label{tab:vlm}
\end{table}

\section{Discussion and Conclusion}
Our Radiologist Copilot demonstrates strong performance in generating CT liver radiology reports. Moreover, it is feasible to generate chest or abdomen radiology reports, as the agent integrates 3D medical VLMs capable of analyzing CT images across different anatomical regions. In addition, the agent leverages LLMs as its reasoning backbone, flexibly selecting appropriate LLMs at runtime based on task complexity and model capability to accomplish tasks efficiently.

In conclusion, Radiologist Copilot is an agentic assistant that autonomously orchestrates specialized tools to address the complex radiology reporting process, going beyond isolated report generation and traditional single-pass methods. By integrating region-of-interest image localization, region analysis planning, strategic template selection, detailed report writing, and rigorous quality control, Radiologist Copilot provides a comprehensive and clinically aligned radiology reporting workflow. Experimental results demonstrate that Radiologist Copilot supports automated and reliable radiology reporting, highlighting its effectiveness in assisting radiologists and enhancing clinical efficiency.

\bibliographystyle{splncs04}
\bibliography{ref}

\end{document}